\documentclass{article}

\usepackage{microtype}
\usepackage{graphicx}
\usepackage{subfigure}
\usepackage{booktabs} % for professional tables
\usepackage{multirow}
\usepackage{parskip}
\usepackage{amsmath,amsfonts,amssymb}

\usepackage{booktabs}
\usepackage{graphicx}

\usepackage{lipsum, babel}
\usepackage{tabularx}

\usepackage{float}
\usepackage{pifont}

\newcommand{\bx}{\mathbf{x}}
\newcommand{\bX}{\mathbf{X}}
\newcommand{\bbf}{\mathbf{f}}

\newtheorem{definition}{Definition}[section]

\usepackage[accepted]{icml2021}

\usepackage{hyperref}

\icmltitlerunning{Submission and Formatting Instructions for ICML 2021}

\begin{document}

\twocolumn[
\icmltitle{Joining datasets via data augmentation in the label space for neural networks}

% It is OKAY to include author information, even for blind
% submissions: the style file will automatically remove it for you
% unless you've provided the [accepted] option to the icml2021
% package.

% List of affiliations: The first argument should be a (short)
% identifier you will use later to specify author affiliations
% Academic affiliations should list Department, University, City, Region, Country
% Industry affiliations should list Company, City, Region, Country

% You can specify symbols, otherwise they are numbered in order.
% Ideally, you should not use this facility. Affiliations will be numbered
% in order of appearance and this is the preferred way.
\icmlsetsymbol{equal}{*}

\begin{icmlauthorlist}
\icmlauthor{Jake Zhao (Junbo)}{equal,zju}
\icmlauthor{Mingfeng Ou}{equal,gr,tj}
\icmlauthor{Linji Xue}{gr}
\icmlauthor{Yunkai Cui}{gr}
\icmlauthor{Sai Wu}{zju}
\icmlauthor{Gang Chen}{zju}
\end{icmlauthorlist}

\icmlaffiliation{zju}{College of Computer Science and Technology, Zhejiang University, Zhejiang, China}
\icmlaffiliation{tj}{Department of Software Engineering, Tongji University, Shanghai, China}
\icmlaffiliation{gr}{Graviti Inc., Shanghai, China}

% \icmlaffiliation{to}{Department of Computation, University of Torontoland, Torontoland, Canada}
% \icmlaffiliation{goo}{Googol ShallowMind, New London, Michigan, USA}

\icmlcorrespondingauthor{Gang Chen}{cg@zju.edu.cn}

% You may provide any keywords that you
% find helpful for describing your paper; these are used to populate
% the "keywords" metadata in the PDF but will not be shown in the document
\icmlkeywords{Deep Learning, ICML}

\vskip 0.3in
]

% this must go after the closing bracket ] following \twocolumn[ ...

% This command actually creates the footnote in the first column
% listing the affiliations and the copyright notice.
% The command takes one argument, which is text to display at the start of the footnote.
% The \icmlEqualContribution command is standard text for equal contribution.
% Remove it (just {}) if you do not need this facility.

%\printAffiliationsAndNotice{}  % leave blank if no need to mention equal contribution
\printAffiliationsAndNotice{\icmlEqualContribution} % otherwise use the standard text.

\begin{abstract}
Most, if not all, modern deep learning systems restrict themselves to a single dataset for neural network training and inference.
In this article, we are interested in systematic ways to join datasets that are made of similar purposes. Unlike previous published works that ubiquitously conduct the dataset joining in the uninterpretable latent vectorial space, the core to our method is an augmentation procedure in the label space. 
The primary challenge to address the label space for dataset joining is the discrepancy between labels: non-overlapping label annotation sets, different labeling granularity or hierarchy and etc. Notably we propose a new technique leveraging artificially created knowledge graph, recurrent neural networks and policy gradient that successfully achieve the dataset joining in the label space. Empirical results on both image and text classification justify the validity of our approach.
\end{abstract}

\section{Introduction}
\label{sec:intro}

The advances of deep learning~\citep{lecun2015deep} arise in many domains, such as computer vision~\citep{krizhevsky2017imagenet}, natural language processing~\citep{NIPS2014_a14ac55a}, speech~\citep{oord2016wavenet}, games~\citep{silver2017mastering} and etc. 
In particular, the most popular paradigm to date is the so-called \emph{end-to-end} learning paradigm. 
Its recipe can normally be summarized as follows: (i)-prepare a dataset consists of numerous \{input-target\} groups; (ii)-feed the dataset to a model, or coupled with a pre-trained model; (iii)-optimizing the model by a gradient-based method and finally, (iv)-inference on testing data points.
%In spite of its massive success, we argue that the data complexity of such paradigm has much room to improve.
In spite of its massive successes, we argue that a given dataset should not be only used once for one phase of task. 
\textbf{Simply put, why train your neural network using just one dataset?}
Instead, its versatility can be substantially enhanced by a novel framework of dataset joining.

In real world applications, we often have the choice of multiple datasets for the same task.
Perhaps some datasets can easily be combined, but some cannot.
The main bottleneck to join datasets together, we argue, is the label discrepancy --- that is, 
the discrepancy caused by inconsistent label set, different semantic hierarchy or granularity.
Prior work attempting to solve this problem has ubiquitously focused on mixing methodologies in the vectorial hidden space, enabled by transfer learning algorithms~\citep{he2019rethinking}, adversarial training~\citep{ganin2016domain} and etc.
These works, however, suffer from a lack of interpretability and oftentimes an inadequate exploitation of the semantic information of the labels.
To the best of our knowledge, how to combine different datasets directly in the label space remains an untouched research challenge.

In this article, we aim to propose a new framework to join datasets and directly address the label space mixing or joining. 
Unlike the conventional deep learning paradigm where for each input the model is optimized towards predicting the corresponding label, our paradigm extends to predicting a trajectory on a \emph{label graph} that traverses through the targeted label node.
The construction of the label graph is the core element in our paradigm. 
Later in this paper, we will describe how to construct it in detail; briefly, the label graph can be perceived as an integration of all the labels (as nodes) from considered datasets, in addition to a set of augmented nodes which are used to bridge the isolated label nodes.

\begin{figure}[ht]
\vskip 0.2in
\begin{center}
\centerline{\includegraphics[width=\columnwidth]{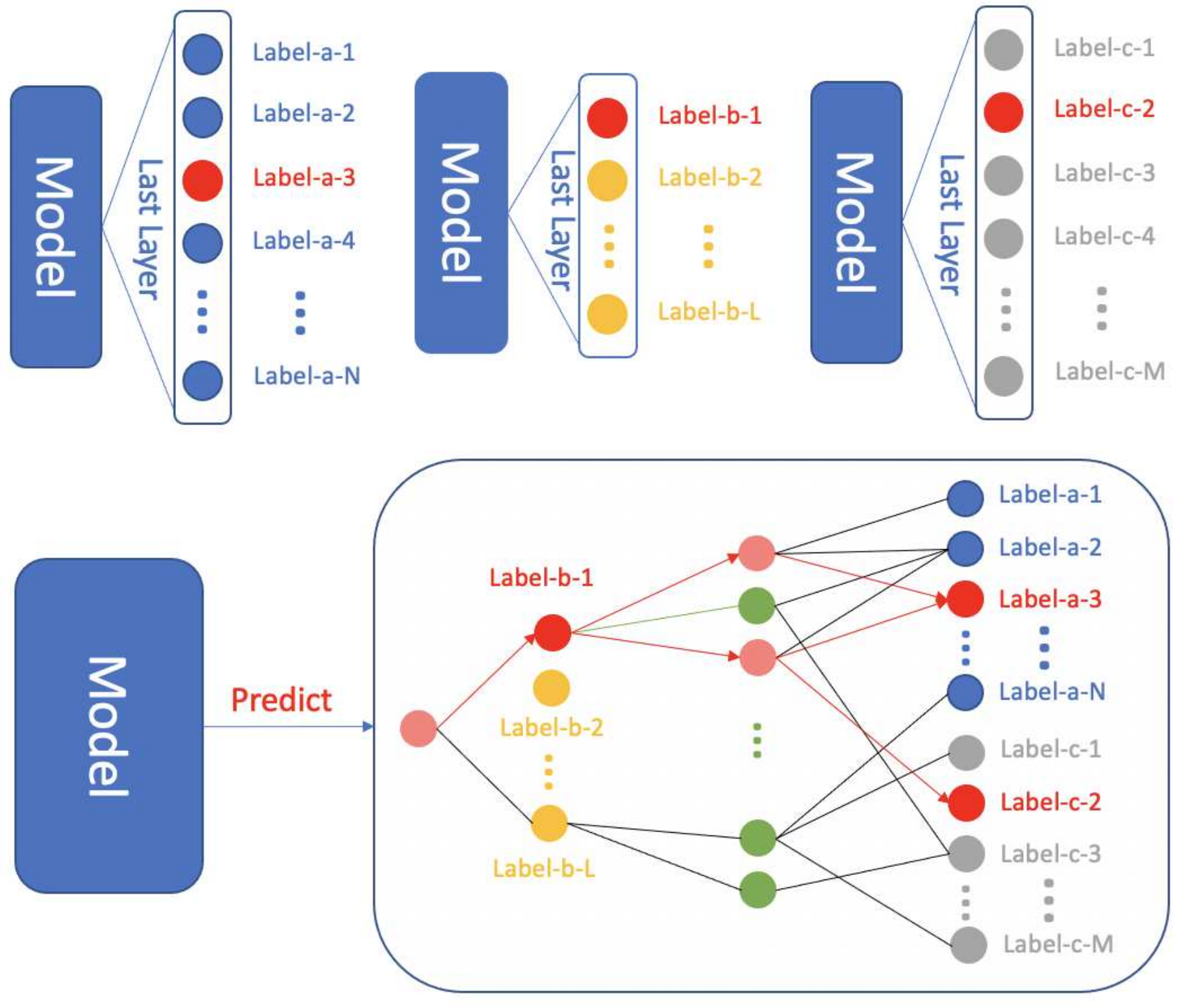}}
\caption{Single label prediction paradigm (top) vs. Label graph prediction paradigm (bottom). Three different colors (i.e., blue, yellow and gray) indicate their source of datasets (blue for dataset-a, yellow for dataset-b and gray for dataset-c). The green and pink ones are the augmented nodes while the red nodes denote the groundtruth targets.}
\label{fig:comparison}
\end{center}
\vskip -0.2in
\end{figure}

In a nutshell, the label graph is knowledge-driven. The construction of it simulates the human process of a decision.
We take cat breed classification as an example. Traditional paradigm may only have delaminated cat breeds label such as $\langle$\emph{british-shorthair}$\rangle$, $\langle$\emph{ragdoll}$\rangle$, and etc. 
By contrast, in our paradigm, the constructed label graph would not only consist of all the cat breed labels as nodes, but also compose several additional \emph{augmented nodes}.
These augmented nodes are functional to indicate \emph{certain features} revealed in its descendant nodes, such as the color feature $\langle$\emph{tabby-color}$\rangle$ being the ancestor node of cat breed label nodes like $\langle$\emph{egyptian-mau}$\rangle$, $\langle$\emph{bengal}$\rangle$ and $\langle$\emph{maine coon}$\rangle$.
Likewise, the augmented nodes can also contain hair features, eye color features and etc.
% functional to categorize  cats, such as  color features \texttt{tabby-color} versus \texttt{single-color}, hair features \texttt{shorthair} versus \texttt{longhair}, eye color features and etc.
% Unlike tranditional softmax-based predictor, our system requires the prediction to ground on an entire trajectory on the label graph that traverses through the groundtruth label node.
An illustration is displayed in Figure~\ref{fig:comparison}.

We postulate our system to have following general merits: (i)-it systematically unifies different label systems from different datasets, tackling the inconsistent label sets;
%This further enables the dataset merging because we could simply form a unified label graph consisting of all label across the datasets. 
(ii)-enhanced transparency and interpretability. Unlike traditional single-label end-to-end prediction being widely criticized as black-box, our paradigm offers a ``decision process'' thanks to the trajectory prediction mechanism.

% We consider the one exemplar scenario. Given two datasets of similar root where the label space of one dataset has rougher granularity than another.
% The goal is to combine the dataset to enhance the performance on both datasets.
% The conventional end-to-end deep learning paradigm fails to do so due to the gap between the label spaces. The solution we intend to promote with this work is described as follows: we first augment the scalar-based labels to a knowledge-driven graph label space $G(V, E)$. The label graph meet the following conditions：
% （i)-all the labels from both datasets can be abstracted as nodes on the graph;
% （ii)-a number of intermediate nodes characterizing the relation between the nodes added as \emph{intermediate nodes} on the graph;
% (iii)-the links \jake{TODO}.
% By contrast to predicting a single label, this paradigm encourages the model to predict one or several trajectories leading to the targeted label sets. 

We instantiate this paradigm by using a recurrent neural network trained with a combination of teacher forcing~\citep{NIPS2014_a14ac55a} and REINFORCE~\citep{williams1992simple} handling different types of augmented nodes.
The contributions of this work are:
\begin{itemize}
\setlength{\itemsep}{0pt}
    \item Propose a new paradigm based on label space augmentation that directly joins datasets in the label space;
    \item Develop a novel instantiation and training algorithm concept-proofing the validity of the paradigm, exhibiting promising results on both image and text classification;
    \item Enhanced interpretability and causal traceability for the proposed approach than the conventional end-to-end paradigm.
\end{itemize}

The rest of the paper is organized as follows. Section~\ref{sec:relatedwork} outlines the related work throughout the literature. In section~\ref{sec:paradigm}, we first formally describe a more abstracted version of the system  and then go to a more concrete model instance. Section~\ref{sec:experiment} shows the experimental results. Section~\ref{sec:conclusion} concludes the paper.

\section{Related Work}
\label{sec:relatedwork}
In this section, we examine the existing related work.

\subsection{Pre-trained models}
The concept of pretrained models has been very popular in recent years. 
From computer vision~\cite{misra2020self,he2019rethinking,chen2020generative}, to natural language processing~\cite{devlin2018bert,yang2019xlnet,radford2019language}, pretrained models demonstrate promising results for performance gain on a variety of domain and tasks.

Fundamentally, the pretrained models are often obtained from a separate and large-scale dataset, and trained under certain criteria (mostly unsupervised). This can be seen as merging the knowledge from this large-scale dataset to the (often much smaller) dataset involved in the downstream task.
Notice that this joining or merging procedure is enacted in the feature space and enabled by gradient-based techniques.

\subsection{Transfer learning}
The idea of transfer learning is to reuse the past experience from a source task to enhance models' performance on a target task, most commonly including parameter finetuning or feature reusing.
This area has been studied for many years~\cite{pratt1991direct,kornblith2019similarity,huh2016makes,yosinski2014transferable,bozinovski2020reminder,yu2017transfer}.
As early as it is in 1976, \citet{bozinovski2020reminder} offers a mathematical and geometrical model of transfer learning in the neural network. Modern work includes \cite{yu2017transfer} investigating transfer learning with noise, \cite{huh2016makes} examining the underlying reason of ImageNet~\cite{ImageNet_VSS09} being a good dataset for transfer learning and \cite{yosinski2014transferable} addressing the transfer capacity at different layers in a neural network.
More recently, \cite{raghu2019rapid} assessed reused feature in the meta learning framework like MAML~\cite{pmlr-v70-finn17a}. \citet{neyshabur2020being} argued that the features from the latter layer from a neural network have a better influence for transfer learning. 
Some negative results are also well conveyed from the community. \cite{kornblith2019better} challenges the generalizability of a pretrained neural network obtained from ImageNet. \cite{he2019rethinking} shows that transfer learning does not necessarily yield performance gains. 

Nonetheless, these works are generally complementary to ours. The primary transfer mechanism is conducted in the feature space, either from parameter porting or gradient-based finetuning. Our paradigm aims directly at the (often discrete) label space joining.

\subsection{Knowledge distillation and pseudo labels}
Recently, \citet{pham2020meta} extends from the original pseudo label idea~\cite{yarowsky1995unsupervised, lee2013pseudo} to reach superior results on ImageNet.
The idea of pseudo label relies on a teacher network to generate \emph{fake} labels to teach a student network.
While we find this line of work similar to ours owing to the label space manipulation, it differs from our approach in two aspects: (i)-prior work on pseudo label is often limited to an unsupervised setup where the pseudo labels are tagged with unlabeled datasets; (ii)-in our paradigm we do not have any machine-generated labels but we rely on a domain knowledge-based label graph.
In addition, we will compare our approach against a revised version of supervised pseudo label in section~\ref{sec:experiment}.

\subsection{Label Relationship and Classification}

Label Relationship has shown promising potential in classification tasks exhibited by existing works.
\citet{deng2014large,ding2015probabilistic} were among the first to point out the concept of label relation graph, in which the authors proposed to construct a label relation graph and used set-based or probabilistic classifiers. However, this line of work cannot deal with nondeterministic paths which commonly exist in an off-the-shelf label graph nor discuss the datasets joining scheme.
\citet{ristin2015categories} adopted  Random Forests and propose a regularized objective function that takes into account unique parent-child relations between categories and subcategories to improve classification by coarser labeled data. \cite{wang2016cnn} proposed a method that recurrently lets the prediction flow through a same set of ground truth labels, so as to exploit co-occurrence dependencies among multi objects in an image and improve multi-classes classification performance. Similarly, \citet{hu2016learning} pointed out visual concepts of an image can be divided into various levels and proposed an rnn based framework to capture inter-level and intra-level label semantics.

Our work is similar to some of the above works mainly from a methodology perspective, but differs in two main aspects: (i) instead of leveraging label relations in a single set(or rather in a single image), our framework is capable of incorporating labels from both internal and external set, and across all hierarchies or granularities; (ii)-our framework can resolve the multi-datasets joining in the label space. 
\section{Method}
\label{sec:paradigm}

% \begin{figure}
%   \centering
%   \includegraphics[width=0.8\linewidth]{framework.eps}\vspace{-2mm}
%   \caption{Our Framework BEIG}
%   \label{fig:framework}
%   \vspace{-3mm}
% \end{figure}

\subsection{Setup}

\subsubsection{Dataset joining problem}
Let $D^{A} = \{(\bx^a_1, y^a_1), (\bx^a_2, y^a_2) \cdots (\bx^a_{N_a}, y^a_{N_a})\}$ and $D^{B} = \{(\bx^b_1, y^b_1), (\bx^b_2, y^b_2) \cdots (\bx^b_{N_b}, y^b_{N_b})\}$ be the targeted datasets respectively.
% We further use $\mathcal{Y}_a$ and $\mathcal{Y}_b$ to denote the label set for the datasets respectively.
The annotated labels for two datasets may or may not overlap.

Conventional deep learning system relies on an end-to-end scheme where two functions are trained separately to map $X$ to $Y$:
$$
f^a(\bX^a) \to Y^a, \quad f^b(\bX^b) \to Y^b.
$$
In this work, we are interested in combining two datasets via a label space joining algorithm. For example, in a classification setup, we intend to overcome the main bottleneck --- the discrepancy between the label sets of the datasets.
The traditional softmax-based classifiers are inefficient for the joining. The foundation of a softmax function is to introduce ``competition'' into the prediction process. The joined label set commonly exists labels that are not mutually exclusive in the taxonomy.

\subsection{Method}

\subsubsection{knowledge-driven label graph}
We consider the most challenging and perhaps most common situation where the label sets have no overlap, $\cap (\mathcal{Y}_a, \mathcal{Y}_b) = \varnothing$.

First and foremost, we construct a label graph $\mathcal{G}(\mathcal{V}, \mathcal{E})$ covering the label sets of both datasets, where $\mathcal{V}$ is the set of $N_v$ nodes, $\mathcal{E} \subseteq \mathcal{V}  \times \mathcal{V}$ is the set of $N_e$ edges connecting $M$ pairs of nodes in $\mathcal{V}$.
Specifically, we construct the graph based on the following steps:
\begin{enumerate}
\setlength{\itemsep}{3pt}
\setlength{\parsep}{0pt}
\setlength{\parskip}{0pt}
    \item start an empty graph with only a root node $v_0$, $\mathcal{V}=\{v_0\}$;
    \item add labels from both datasets as nodes to the graph,
    $\mathcal{V}=\{v_0\} \cup \mathcal{Y}^a \cup \mathcal{Y}^b$;
    \item based on the labels' semantic information and domain knowledge, we add augmented nodes $\mathcal{Y}^{\text{aug}}$ to the graph, $\mathcal{V}=\{v_0\} \cup \mathcal{Y}^a \cup \mathcal{Y}^b \cup \mathcal{Y}^{\text{aug}}$;
    \item add links $\mathcal{E}$ to connect the nodes;
\end{enumerate}
Oftentimes, when constructing the label graphs, we primarily want to base it on the domain knowledge that is accumulated across several decades for the considered task. Take the pet breed classification as an example. We tweaked slightly the above steps into: (i)-we crawl a complete knowledge graph $K$ from a domain website like Purina; (ii)-we start with an empty graph $G$ and place the root node $\langle$\emph{animal}$\rangle$; (iii)-going top-down the taxonomy of $K$, we select related nodes $\langle$\emph{cat}$\rangle$ and $\langle$\emph{dog}$\rangle$ and place them under the root; (iv)-repeat (iii), going further down through $K$, we place the augmented nodes in $K$ representing the visual features describing a pet like hair, ear, tail features, etc. (v)-repeat the previous steps until the finest granularity of $K$. 

\begin{figure}[t]
\vskip 0.2in
\begin{center}
\centerline{\includegraphics[width=\columnwidth]{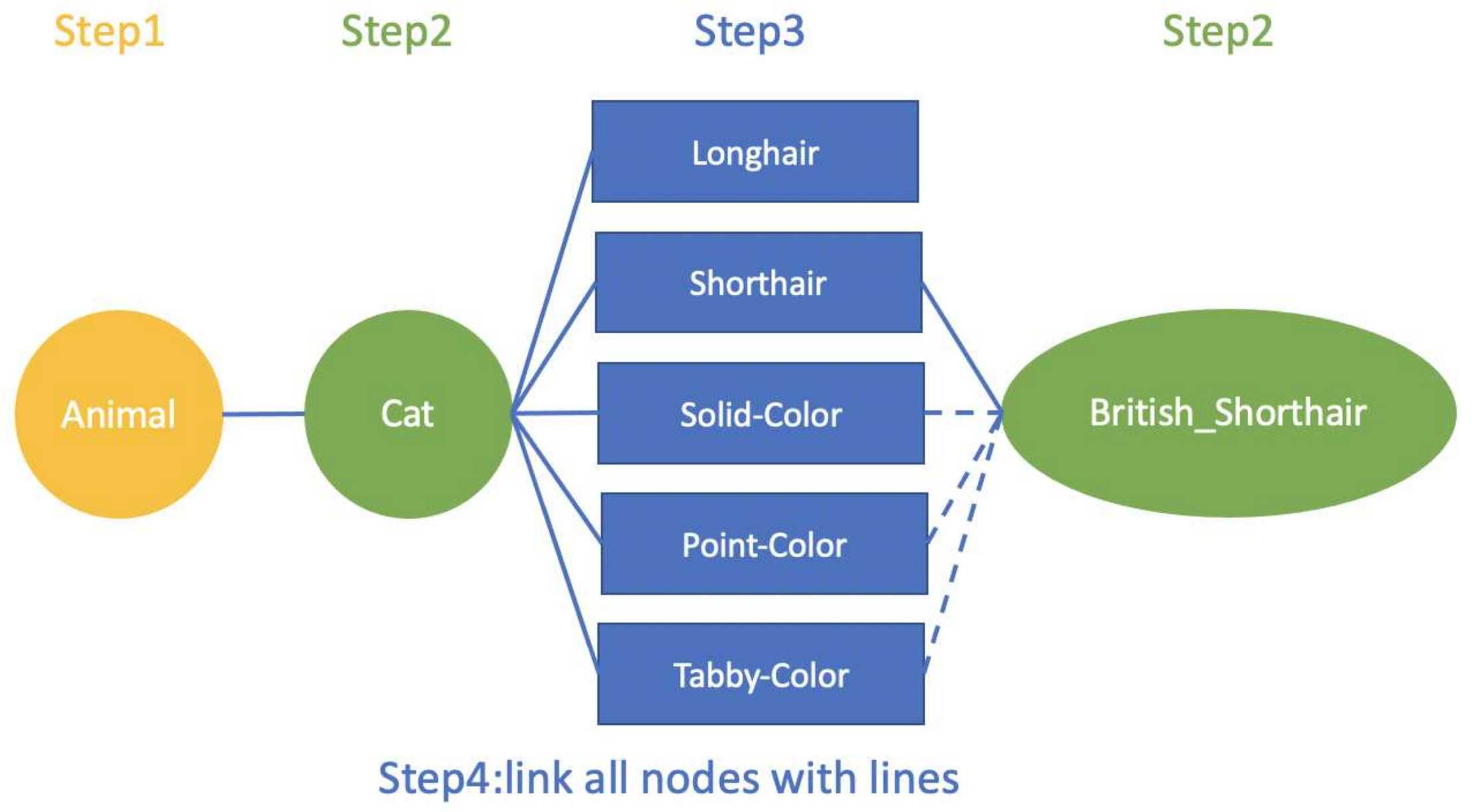}}
\caption{A subset example of constructed Label Graph. The root node is colored yellow. Two label nodes from $\mathcal{Y}^a$ and $\mathcal{Y}^b$ are colored green. Augmented nodes in $\mathcal{Y}^{\text{aug}}$ are blue.). The dashed line indicates nondeterministic paths, see text for more details.}
\label{fig:labelgraph}
\end{center}
\vskip -0.2in
\end{figure}

The exemplar label subgraph is plotted in Figure~\ref{fig:labelgraph}.
% As we can see, the graphs consists of two different types of nodes, the original nodes drawn from $\mathcal{Y}_a$ and $\mathcal{Y}_b$, and the augmented ones based on the domain knowledge from $\mathcal{Y}_i$.

Note that the label graphs are inherently obtained from the domain knowledge which has been accumulated for decades across different fields. In particular, the label graphs we used for the experiments are all off-the-shelf with minor filtering and modification. For instance, we got the Pet label graph from websites like Purina \footnote{https://www.purina.com/}, and the Arxiv label graphs are collected by its website routing logics. See section~\ref{sec:experiment} for more details.
% Furthermore, we also tend to position our work as a preliminary attempt to incorporate the rich domain knowledge (formatted as knowledge graphs) to boost the connectivism (such as a neural network classifier).

In terms of the \textbf{scalability} or \textbf{expandability} of our approach, we argue that extending a label graph is much cheaper than enlarging the annotated dataset itself because the latter requires significantly more human efforts in annotation.

\subsubsection{Prediction paths}
In this subsection, we define \emph{prediction paths}.
Unlike the conventional deep learning paradigm maximizing the log-likelihood of the predicted label conditioned on its input, our system optimizes the log-likelihood of an entire prediction path (or trajectory) that traverses from the root node to the groundtruth label node.

More formally, we define a prediction path $P$ obtained by running a graph traversal algorithm: $v_0 \to v_1 \to v_2 \to v_{\left<\text{eop}\right>}$, where $v_0$ is the root node and ``eop'' indicates the end of a path. In the remainder of the paper, we may abbreviate this form to $v_0 \circ v_1 \circ v_2 \circ v_3 \circ v_{\left<\text{eop}\right>}$.
As we mentioned in the previous subsection, some of the nodes on $P$ might be a label node belonging to $\mathcal{Y}^a$ or $\mathcal{Y}^b$ while some of the others might be augmented ones $\mathcal{Y}^{\text{aug}}$.
In addition, when one or several nodes on the path are label nodes, we may also write this prediction path down as $P(v_j)$, when $v_j$ is a label node.

Let us give a concrete example. Given a data point sampled from either datasets, $(x, y)$.
We first locate the label node $v$ representing $y$ on the label graph $\mathcal{G}$, then we run a graph traversal algorithm to obtain all possible prediction paths traversing from $v_0$ to $v$. This process winds up with a set of prediction paths, $(P_0(v), P_1(v), \cdots P_{M}(v))$.
The loss objective in the system is to maximize the log-probability of all collected prediction paths.
In the remainder of the paper, we may also call these paths as groundtruth prediction paths, or in short groundtruth paths. 

% Nevertheless in our paradigm, the system is optimized by predicting an \emph{entire path / trajectory} which traverses from the root node to the considered label node.

% More formally, given a data point from either of these datasets, $(x_i, y_i)$, with the label $y_i$ corresponds to the node $v_i \in \mathcal{V}$ on the graph. we run a graph traversal algorithm to get all paths from the root node traversing to $v_i$, denoted as $\mathbf{P}_i \in \mathcal{P}$.
% The loss functions in our paradigm are built to maximize the log-probability of these groundtruth paths.

\subsubsection{Paths categorization}
Briefly we categorize the groundtruth paths into two types: the \emph{deterministic} and \emph{nondeterministic} path.

\begin{definition}[Competing nodes]
We define $u$ and $w$ are competing nodes when $u$ and $w$ share a same ancestor node, and they are mutually exclusive in the taxonomy.
\end{definition}

% certain paths
\begin{definition}[Deterministic path]
Given a path anchoring label node $v$, $P(v)$, the path is deterministic if there does \textbf{not} exist another path $P'(v)$ such that the following conditions are met: node $u \in P(v)$, node $w \in P'(v)$, $u$ and $w$ are competing nodes, and $u \ne w$.
\end{definition}

\begin{definition}[Nondeterministic path]
Given a path anchoring label node $v$, $P(v)$, the path is nondeterministic if there does exist another path $P'(v)$ such that the following conditions are met: node $u \in P(v)$, node $w \in P'(v)$, $u$ and $w$ are competing nodes, and $u \ne w$.
\end{definition}

Let us give a concrete example. In Figure~\ref{fig:labelgraph}, there are two label nodes colored green and five intermediate nodes colored blue. Among these intermediate nodes, $\langle$\emph{Shorthair}$\rangle$ and $\langle$\emph{Longhair}$\rangle$ are competing nodes which can be named as \emph{Hair}  group, and every two nodes among $\langle$\emph{Solid-Color}$\rangle$, $\langle$\emph{Tabby-Color}$\rangle$ and $\langle$\emph{Point-Color}$\rangle$ are also competing nodes which can be named as \emph{Color Pattern} group. Thus, these nodes form one deterministic path (\emph{Animal}$\circ$\emph{Cat}$\circ$\emph{Shorthair}$\circ$\emph{British-Shorthair}) because all samples of $\langle$\emph{British-Shorthair}$\rangle$ is short hair, and three nondeterministic paths (e.g., \emph{Animal}$\circ$\emph{Cat}$\circ$\emph{Solid-Color}$\circ$\emph{British-Shorthair}) because the color pattern of $\langle$\emph{British-Shorthair}$\rangle$ samples can be any node in \emph{Color Pattern} group.

It is relatively straightforward to tackle the prediction of the deterministic paths. It can be implemented by a standard seq2seq~\citep{NIPS2014_a14ac55a} model trained by maximum log-likelihood estimation (MLE) assisted by the teacher forcing mechanism. Note that, different groups of competing nodes use softmax independently because we assume that nodes in different groups are independent.
On the other hand, when facing nondeterministic paths, teacher forcing cannot be applied due to the uncertainty from the intermediate nodes. Therefore, we use the policy gradient algorithm to estimate the gradient from the nondeterministic paths.

\subsubsection{Block softmax}
For general softmax, its output represents the relative probability between different categories, which means that all the categories are competing against each other. However, in our architecture, the competitive relationship only exists among competing nodes. Thus, in order to handle this situation, we adapted softmax as follows: 
$$
p_{u}=\frac{e^{z_{u}}}{\sum_{w \in \mathbb{S}_{u}} e^{z_{w}}}
$$
where $\mathbb{S}_{u}$ denotes the set of competing nodes of node $u$, and $z$ denotes the input to the softmax.
This means that we limit the calculation of relative probability within each competing node groups.
We call this revised version as block softmax function.

More concretely, for the Pet problem, there are nodes describing hair feature and color pattern feature respectively. The nodes inside each of these two groups (or blocks) are \emph{competing nodes}. Softmax inherently introduces competition inside considered predicted labels, hence we place two softmax classifiers for these two groups (or blocks) and the nodes across the groups do not influence each other. 

\subsubsection{Deterministic path prediction}
Essentially we treat each of the groundtruth deterministic paths as a sequence and let a decoder predict each token (e.g. node) autoregressively.
Because every token on a deterministic path is guaranteed to be the groundtruth label against their competing nodes respectively, this allows us to borrow the flourishing literature from the sequence decoding~\citep{NIPS2014_a14ac55a,cho2014learning,DBLP:journals/corr/ChungGCB14}. 
That being said, we adopt teacher forcing as our training standard.

Formally, given a deterministic path $P = (v_0, v_1, v_2, ... v_N)$, we feed the sequence into a recurrent unit and adopt the teacher-forcing strategy~\cite{doi:10.1162/neco.1989.1.2.270} during training.
Therefore we can gather the hidden state as follows:
\begin{equation}
    \mathbf{f}_t = g(\mathbf{e}_{t}, \mathbf{f}_{t-1}),
\end{equation}
where $g$ is a compositional function such as gated recurrent units~\cite{cho2014learning}, $\mathbf{f_t}$ denotes the feature vector at the token step $t$, $\mathbf{e}$ is a learned node embedding matrix.

At the top layer of the recurrent unit, we extract the feature $\mathbf{f}$ and then carry it to a block softmax predictor at each position.
The recurrent decoder maximizes the conditional log-probability $\log p( v_0, v_1, v_2 \cdots v_{\left<\text{eop}\right>})\ |\ \bx)$
Note this training process is analogous to many natural language processing tasks, such as machine translation~\cite{NIPS2017_3f5ee243}. At each step $t$, the overall objective for the certain paths can be derived in an autoregressive fashion:
\begin{equation}
\label{eqa:certain}
    \mathcal{L}_d = - \sum_t \log p(v_{t+1} | v_t, \mathbf{f}_t)
\end{equation}
The gradient from predicting the deterministic paths can be easily and tractably computed using backpropagation.

\subsubsection{Nondeterministic paths}
Predicting nondeterministic paths is generally more challenging, due to the nondeterministic nodes in the groundtruth path blocking the usage of normal sequence training techniques like teacher forcing.
To cope with this, we first define a reward function.
\begin{equation}
    r(\hat{P}) = \frac{1}{|S|} |\hat{P} \cap S|,
\end{equation}
where $\hat{P}$ is a generated path sampled by the model (from a multinomial distribution produced at each node step),  $S$ is a set composed by the groundtruth label nodes corresponding to the input, $|\cdot|$ denotes the size of a set.

Then we write down the loss function for the nondeterministic path prediction:
\begin{align}
    \mathcal{L}_{nd} = - \sum \log p(\hat{P})r(\hat{P}) = -\mathbb{E}_{\hat{P} \sim p(P)} r(\hat{P}),
\end{align}

where $p(P)$ is the path prediction produced by the model and the $r(\hat{P})$ is the associated reward.
In practice, we approximate the expectation via a single sample generated from the distribution of action (node choices).

Without too much details, the gradient of $L_{nd}$ can be estimated by:
\begin{equation}
    \frac{\partial \mathcal{L}_{nd}}{\partial \bbf_t} = (r(\hat{P}) - b(r)) p(\hat{v}_{t+1} | \hat{v}_t, \bbf_t),
\end{equation}
where $b(r)$ is an average baseline estimator.
For simplicity we omit the details of the gradient derivation, we refer the readers to \citet{williams1992simple} for more details.

% For simplicity we omit the gradient derivation and we refer the readers to ~\cite{} for details.
% To compute the gradient from the nondeterminstic nodes on the path, we have to rely on the Policy Gradient estimation~\cite{}:
% \begin{equation}
    
% \end{equation}
% Empirically we find this whole-sequence reward estimation works quite well, without the need for partial sequence estimation like ~\citet{}.

% With slight abuse of notations, we consider a set of nondeterministic path $P = (v_0, v_1, \cdots [v_t^0, v_t^1, \cdots v_t^{N_u}], \cdots v_N)$ where $N_u$ dubs the total number of uncertain nodes at stage $t$.
% The main obstacle to apply ordinary sequence training method and backprop is due to the uncertain nodes and their discrete nature as in $[v_t^0, v_t^1, \cdots v_t^{N_u}]$.
% To compute the gradient from these uncertain nodes, we have to rely on Policy Gradient estimation~\cite{}. 
% Namely,
% \begin{equation}
%     \nabla \mathcal{L}_{\text{uncertain}} = \nabla [\log P(T | v_t', \mathbf{F}) - b(v_t)],
% \end{equation}
% \jake{path changed to trajectory}
% where $T$ denotes the groundtruth node path, $v_t'$ is a sampled result from the multinomial distribution parametrized by $P(v_t | v_{t-1} \cdots v_{0}, x)$,
% $b(v_t)$ is a baseline estimator,
% $P(T | v_t')$ is factorized the same way as in equation~\ref{eqa:certain}, and we estimate it based on the certain nodes on the given trajectory.

\subsubsection{Model instantiation}
To this end, we finalize the instantiation of a model instance fitting in this paradigm.
As we described earlier, our model is devised to predict a node path rather than a single node. 
%This paradigm is, in theory, capable of handling both certain and uncertain types of node paths.

\begin{figure*}[t]
\vskip 0.1in
\begin{center}
\centerline{\includegraphics[width=1.8\columnwidth]{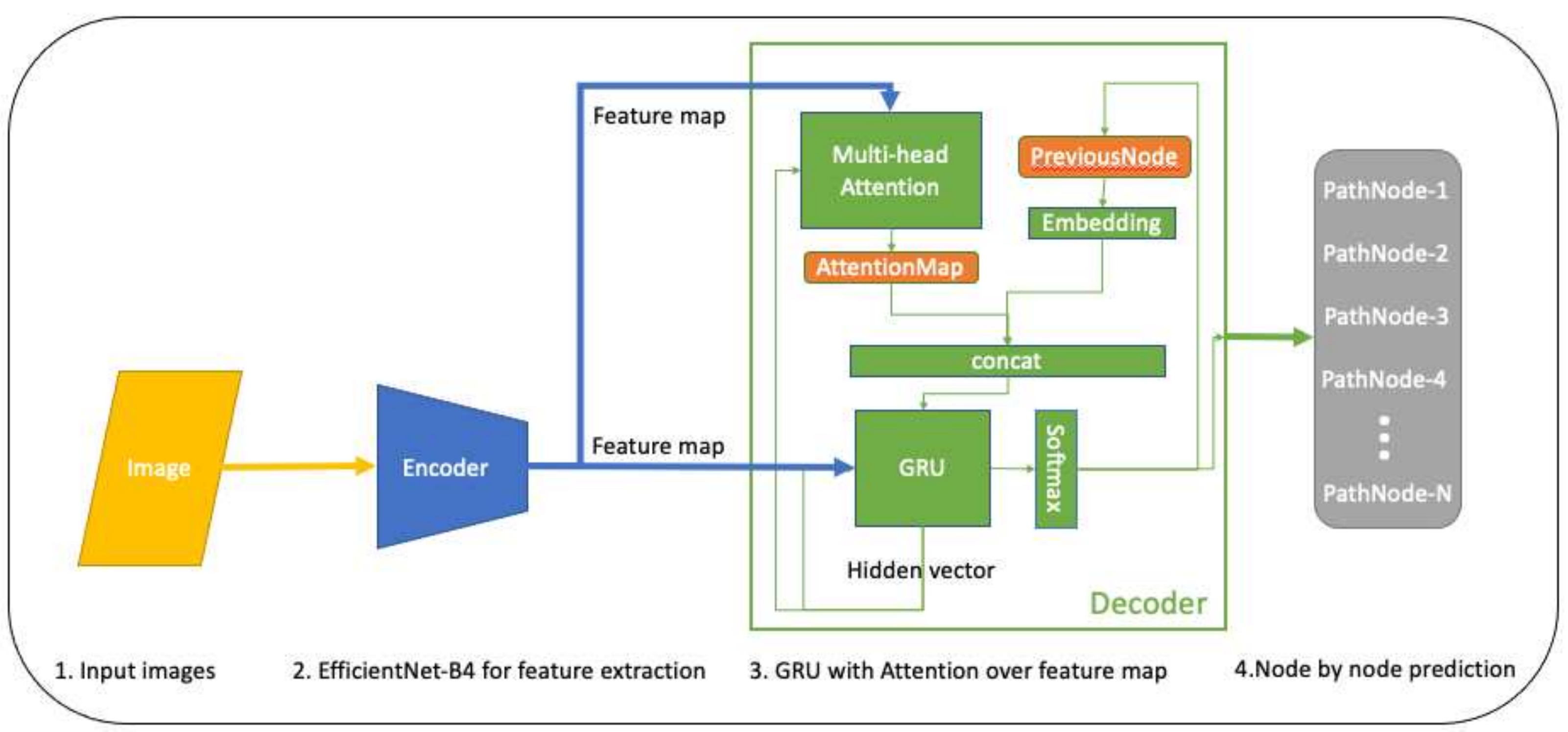}}
\caption{Our model architecture}
\label{fig:modelarch}
\end{center}
\vskip -0.1in
\end{figure*}

The overall model instantiation in computer vision tasks is illustrated in Figure~\ref{fig:modelarch}.
Briefly, this model employs a seq2seq alike structure~\cite{NIPS2014_a14ac55a};
we use an EfficientNet-B4~\cite{tan2019efficientnet} as our encoder backbone, and a gated recurrent unit in the decoder.
As we described earlier, the overall model is trained by a combination of techniques: gradient descent, backpropagation, teacher forcing and policy gradient. Our training procedure is compiled Algorithm~\ref{alg:tranalg}. 
Meanwhile, during the inference phase, we simply apply a greedy decoding algorithm, showed in Algorithm~\ref{alg:algorithm}.
Analogously, in natural language processing task, we simply replace the EfficientNet structure with BERT~\cite{devlin2018bert}.

% \subsubsection{Decoding strategy}

% the decoding algorithm is compiled in , which is a kind of greedy algorithm. Note that, in \emph{Decoder}'s inference phase, $LabelGraph$ is used to generate mask to limit predictable node set according to previous predicted node.

\begin{algorithm}[tbp]
\caption{Training algorithm}
\begin{small}
\label{alg:tranalg}
    \textbf{Input}: images $\{x^{(k)}\}^{m}_{k=1}$, paths $\{P^{(k)}\}^{m}_{k=1}$ ($P^{(k)} = \{v^{(k)}_{1}, ..., v^{(k)}_{n}\}$), uncertain sample indexes $I_{pg}=\{a_{k^{'}}\}_{k^{'}=1}^{q}$($0 \leq q \leq m, 1 \leq a_{k^{'}} \leq m$ )\\
    
    \textbf{Parameter}: LabelGraph $\mathcal{G}$, MaxLength $n$, Teacher Force Rate $r_{tf}$($0 \leq r_{tf} \leq 1$)\\
    \begin{algorithmic}
    \STATE Let loss $\mathcal{L} = 0$
    \STATE \textbf{/*1.Train deterministic path by teacher forcing*/}
    \STATE Let Sample $coin \sim \mathcal{U} (0,1) $ 
    \STATE Let token $t_{0} = {\{ {\langle START \rangle}^{(k)}\}}^{m}_{k=1} $
    \STATE Define encoder function $g_{\text{enc}}$, decoder function $g_{\text{dec}}$
    \STATE Compute feature $f = g_{\text{enc}}(\{x^{(k)}\}^{m}_{k=1})$
    \STATE Initial $\mathbf{f}_0$ by $f$
    \FOR{$i=1$ {\bfseries to} $n$}
        \STATE Let probabilities $p_{i}, \mathbf{f}_i = g_{\text{dec}}(\mathcal{G}, f, \mathbf{f}_{i-1}, t_{i-1})$
        \STATE Compute loss at time step $i$ and accumulate: $\mathcal{L} = \mathcal{L} + (-\log p(t_{i} | t_{i-1}, \mathbf{f}_i))$
        \IF{$coin \leq r_{tf}$}
            \STATE  $t_{i} = \{v^{(k)}_{i}\}^{m}_{k=1}$
        \ELSE
            \STATE $t_{i} = $ Indexes of max values in $p_{i}$
        \ENDIF
    \ENDFOR
    \STATE
    \STATE  \textbf{/*2.Train nondeterministic path by policy gradient*/}
    \IF {$I_{pg}\ is\ not\ empty$ }
        \STATE Sample $\{x^{(k)}\}^{m}_{k=1}$ according to $I_{pg}$ and assign to $x^{pg}$
        \STATE Sample $\{P^{(k)}\}^{m}_{k=1}$ according to $I_{pg}$ and assign to $P^{pg}$
        \STATE Let $t_{0} ={\{{\langle START \rangle}^{(k^{'})}\}}^{q}_{k^{'}=1}$
        \STATE Compute feature $f = g_{\text{enc}}(x^{pg})$
        \STATE Let path probabilities $pp = [\ ]$
        \FOR{$j=1$ {\bfseries to} $n$}
            \STATE Let probabilities $p_{j} = g_{\text{dec}}(\mathcal{G}, f, t_{j-1})$
            \STATE Sample nodes $t_{j} \sim p_{j}$
            \STATE Append sampled nodes' probabilities $p^{t_{j}}_{j}$ at time step $j$ to $pp$
        \ENDFOR
        \STATE Compute loss $\mathcal{L} = \alpha*\mathcal{L} + \beta*(- \sum \log pp*r(t_{n-1}))$
    \ENDIF
    \STATE Backpropagate $\mathcal{L}$, and update $g_{\text{enc}}$, $g_{\text{dec}}$.
\end{algorithmic}
\end{small}
\end{algorithm}

\begin{algorithm}[tbp]
\caption{Decoding algorithm}
\begin{small}
\label{alg:algorithm}
    \textbf{Input}: image $x$\\
    \textbf{Parameter}: LabelGraph $\mathcal{G}$ , MaxLength $n$\\
    \textbf{Output}: Path based on LabelGraph $\mathcal{G}$
\begin{algorithmic}
    \STATE Let token $t_{1} =\langle START \rangle$
    \STATE Let path $p=[\ ]$
    \STATE Define encoder function $g_{\text{enc}}$, decoder function $g_{\text{dec}}$
    \STATE Compute feature $f = g_{\text{enc}}(x)$
    \FOR{$i=1$ {\bfseries to} $n$}
        \STATE Let $prob = g_{\text{dec}}(\mathcal{G}, f, t_{i})$
        \STATE $t_{i} = $ Index of Max value in $prob$
        \IF{ $t_{i} == \langle EOP \rangle$}
            \STATE break
        \ELSE
            \STATE Append $t_{i}$ to $p$
        \ENDIF
    \ENDFOR
    \STATE \textbf{return} $p$
\end{algorithmic}
\end{small}
\end{algorithm}

\section{Experimental Evaluation}
\label{sec:experiment}
In this section, we show the empirical results using our presented architecture.

\subsection{Setup}

\textbf{Datasets}
To validate our approaches, we experiment with two modalities of data: images and natural language.
And just for proof-of-concept, we choose the most conventional task --- image and text classifications. The datasets statistics are shown in Table \ref{tab:dataset}. 
In the domain of computer vision, we use (i)-the Oxford-IIIT Pet dataset~\cite{parkhi12a} and Dogs vs. Cats\footnote{https://www.kaggle.com/c/dogs-vs-cats} as a group; (ii)-102 Category Flowers dataset~\cite{Nilsback08} and the 17 Category Flowers dataset~\cite{Nilsback06} as another group.
In particular, the chosen dataset groups can be characterized as having one dataset very fine-grained annotated with the other being much coarser. 
To expand the horizon of the experiments, in group (i) there is \textbf{no} label overlap between the datasets, while for the group (ii) \textbf{8} labels are shared in the sets.
Notice that, for evaluation purposes, we aim at the performance on the finer-grained datasets. It is relatively easy to enhance the coarser-level performance by fusing the finer-grained labels (through a label set mapping table for example).
So we focus on the significantly more challenging task where we intend to enhance the performance on finer-grained datasets by utilizing its coarsely annotated counterpart.

On the other line of natural language processing experiments, we use the Arxiv dataset~\cite{clement2019arxiv}.
The downloadable version of this dataset involves a gigantic number of articles released on arxiv throughout the entire 28 years. 
In this work, we only take a subset to conduct our experiments.
The goal for this set of experiments is for a hierarchical multi-label text classification, thanks to the naturally hierarchical labels from Arxiv\footnote{https://arxiv.org/}, such as "cs.machine-learning".
In particular, we artificially prepare two datasets: an \emph{Arxiv Original} dataset with 50,000 samples and an \emph{Arxiv Augment} dataset with 50,000 samples but only maintaining the coarser-level labels (such as "cs").
Similar to computer vision tasks, we conduct the experiments on both the \emph{Arxiv Original} dataset and the combined dataset dubbed \emph{Arxiv Fusion}. 

% \textbf{Datasets.} We use two groups of public datasets as shown in Table~\ref{tab:dataset}, where $K$ denotes the number classes and $l$ represents the rough sample quantity for each class. Note that, \emph{PetFusion} dataset's train data is obtained by merging \emph{Oxford-IIIT Pet}'s train data with \emph{Dogs vs. Cats}'s whole data, and it's test data is the same as \emph{Oxford-IIIT Pet}'s. For the other group, \emph{FlowerFusion} dataset's train data is obtained by merging \emph{102 Category Flower}'s train data with \emph{17 Category Flower}'s data with same labels.   
% \emph{Oxford-IIIT Pet} is employed for fine grained object categorization of the breeds of cat and dog. \emph{Dogs vs. Cats} dataset is used for a competition on Kaggle \footnote{https://www.kaggle.com/c/dogs-vs-cats/rules.}. \emph{102 Category Flower} dataset and \emph{17 Category Flower} dateset are used for fine grained classification for the flower species, and both of them are provided by VGG of University of Oxford.

\textbf{Baseline setup}
In computer vision tasks, we adopt the state-of-the-art model EfficientNet~\citep{tan2019efficientnet} as the baseline results that we compare against.
The reasons are the following:
(i)-EfficientNet framework is one of the state-of-the-art pretrained frameworks in computer vision. Its performance on the Pet datasets is ranked the top on the board.
(ii)-As we directly use a pretrained EfficientNet framework as our encoder, it is very straightforward to see if our techniques of label augmentation and dataset fusion are effective.
We use EfficientNet-B4 version instead of their biggest B7 version due to time and space complexity issues.

In addition, we compare against the pseudo label method~\cite{lee2013pseudo}. 
Take the pet group datasets as an example. We first train a standard fine-grained classification model, and use it to generate \emph{pseudo labels} on the images provided by the coarser dataset, i.e. Dogs vs Cats. We additionally use the datasets' coarse labels (dogs versus cats) to conduct a filtering (if the produced pseudo label is incorrect and detected by the filter, we remove the data sample from the fused set). Thereby we have a combined dataset that possesses a unified fine-grained label system.
We argue that this revision improves from the original unsupervised version of pseudo label~\cite{lee2013pseudo} because we further leverage some weak supervision derived from the coarsely annotated labels.

Besides the above baseline setups,  we conduct a multi-label classification setting dubbed as \emph{Label Set}, adopted from one of the pivotal experiments in \citet{deng2014large}.
Specifically, we flatten all the labels on a ground-truth path into a set of labels and train a set-predictive multi-label classification network. Essentially this setup contains the same volume of information as our approach but lacking the label taxonomy information. We hope to use this comparison to demonstrate the necessity of the label graph.

% Specifically, we take the labels on the ground-truth path for a sample as its ground truth label sets for classification. This setting provides the same amount of supervision information in aspects of labels as ours, but lacks label graph.

Notably, the natural language processing experiments are established by two settings: one with a pretrained BERT model and the other one adopts an LSTM-based encoder trained on-the-fly.

\subsection{Main results}
We report our results in Table~\ref{tab:baseline} and Table~\ref{tab:text_result} respectively for vision and text classification.
We may conclude from the scores that:
(i)-the augmentation strategy in the label space enables dataset joining. The models we obtained from the fused dataset perform substantially better than the end-to-end learning paradigm.
It also performs better than the improved version of the weakly-supervised pseudo label method;
(ii)-Even without the help from the extra dataset, simply augmenting the label into label graph coupling with our training strategy still offers certain performance gains.
We reveal the experimental details in the Appendix.

% For image classification task, we choose a state-of-the-art model EfficientNet\citep{tan2019efficientnet} provided by Google to build baseline and be adopted as backbone for our architecture. And we choose  \emph{EfficientNet-B4} with 19M parameters instead of  \emph{EfficientNet-B7} with 66M parameters on many image classification benchmarks for training cost and time reasons. 
% To be specific, we fine-tune \emph{EfficientNet-B4} to obtain two types of baselines for single label and multi labels classification respectively, as shown in Table~\ref{tab:baseline}. For \emph{EfficientNet-B4(Multi-Label)}, we split the nodes on ground truth path into multiple labels so that it can be transferred to multi-label prediction task. Note that, the \emph{PetFusion} dataset for \emph{EfficientNet-B4(Single Label)} is obtained by merging high-probability pseudo labels into \emph{Oxford-IIIT Pet}, and pseudo labels is generated on \emph{Dogs vs. Cats} dataset by the model \emph{EfficientNet-B4(Single Label)} that trained on \emph{Oxford-IIIT Pet}.    

\begin{table}[t]
\caption{Dataset statistics. $K$ denotes the number of classes. Note that the testing data for datasets with coarser annotation is \textbf{not} used.}
\label{tab:dataset}
\vskip 0.15in
\begin{center}
\begin{small}
\begin{sc}
\begin{tabular}{lrrr}
\toprule
\textbf{Dataset}  & \textit{\textbf{\#Train Data}} & \textit{\textbf{\#Test Data}} & \textit{\textbf{K}}\\ 
\midrule
\textit{Oxford-IIIT Pet} & 3680   & 3669   & 37        \\ 
\textit{Dogs vs. Cats}     & 12,500   & -  & 2       \\ 
\textit{PetFusion}     & 28,680   & 3669   & 39    \\ 
\midrule
\textit{102 Category Flower}     & 1020   & 6149   & 102   \\  
\textit{17 Category Flower}     & 680   & -   & 17  \\  
\textit{FlowerFusion}     & 1180   & 6149   & 102    \\ 
\midrule \midrule
\textit{Arxiv original} & 50000 & 50000 & 149   \\
\textit{Arxiv augment} &  50000 & -  &  21 \\
\textit{Arxiv fusion} &  50000 & 50000  &   149 \\
\bottomrule
\end{tabular}
\end{sc}
\end{small}
\end{center}
\vskip -0.1in
\end{table}

% \subsection{Experimental Setup}

% \jake{TODO move this down.}In the first stage that building baselines, we use two settings: i) batch size 16; initial learning rate 5e-5 that decays by 0.5 every 15 epoch. ii) batch size 12; initial learning rate 1e-4 with dynamic reducing that learning rate would decay by 0.5 if validation accuracy doesn't improve for 5 epochs.The setting i is employed for \emph{Cat-Dog-Classification}, and the other one is for \emph{Flower-Classification}. And in the second stage that training our models, we use the setting: batch size 12; initial learning rate 5e-5 for encoder and 4e-4 for decoder with dynamic reducing that learning rate would decay by 0.5 if validation accuracy doesn't improve for 5 epochs; teacher force rate 0.8. And we use Adam with $\beta_1$ = 0.9 and $\beta_2$ = 0.999. 

% \textbf{Performance Metrics.} In order to verify the performance of image classification methods, popular performance metric (accuracy) is used. All experiments were implemented in Python 3.8 on Pytorch 1.5.1. The experiments were conducted on an Intel(R) Core(TM) i9-9900X CPU 3.50GHz processors (10 physical cores and 20 CPU threads) with 128GB RAM accelerated by a NVIDIA GeForce RTX 2080 Ti GPU.

\begin{table}[t]
\caption{Text classification results, reported in F1.}
\label{tab:text_result}
\vskip 0.15in
\begin{center}
\begin{small}
\begin{sc}
\begin{tabular}{lrrr}
\toprule
\textbf{Model}   & \textit{\textbf{Arxiv original}} & \textit{\textbf{Arxiv Fusion}} \\ 
\midrule
 \textit{BERT + FFN}    & 75.5   &  76.9 \\ 
 \textbf{\textit{BERT + label-aug (ours)}} & 77.4 &\textbf{ 79.3 }\\
\midrule
 \textit{LSTM + FFN}    & 69.9   &  $-$  \\ 
 \textbf{\textit{LSTM + label-aug (ours)}} & \textbf{72.4} & $-$  \\
\bottomrule
\end{tabular}
\end{sc}
\end{small}
\end{center}
\vskip -0.1in
\end{table}

\begin{table*}[t]
\caption{Image classification results reported in accuracy (\%). These results are obtained on the finer-grained testing datasets. (``\ding{53}'' denotes the considered model cannot be directly employed for the setup while ``$-$'' indicates the experiment is less prioritized.)}
\label{tab:baseline}
\vskip 0.15in
\begin{center}
\begin{small}
\begin{sc}
\begin{tabular}{lrrrr}
\toprule
\textbf{Model}   & \textit{\textbf{Oxford-IIIT Pet}} & \textit{\textbf{PetFusion}} & \textit{\textbf{102 Category Flower}} & \textit{\textbf{FlowerFusion}} \\ \midrule
 \textit{EfficientNet-B4~\cite{tan2019efficientnet} + FFN} & 93.84   & \ding{53}  & 90.91 & 91.01     \\  
 \textit{EfficientNet-B4 + Label Set~\cite{deng2014large}}   & 92.58   & $-$  & 85.28 & $-$\\ 
 \textit{EfficientNet-B4 + Pseudo Labels}    &   \ding{53}  & 91.50  & \ding{53}  & \ding{53} \\ 
\midrule
 \textit{\textbf{EfficientNet-B4 + label-aug (ours)}}   & \textbf{94.66}   & \textbf{94.95} & \textbf{92.80} & \textbf{93.27}  \\ 
\bottomrule
\end{tabular}
\end{sc}
\end{small}
\end{center}
\vskip -0.1in
\end{table*}
%  \textit{\textbf{EfficientNet-B4 + label-aug (ours) + PG}}   & \textbf{93.95}   & $-$ & $-$ & $-$ \\

\subsection{Interpretability results}
\begin{figure*}[t]
\vskip 0.1in
\begin{center}
\centerline{\includegraphics[width=1.9\columnwidth]{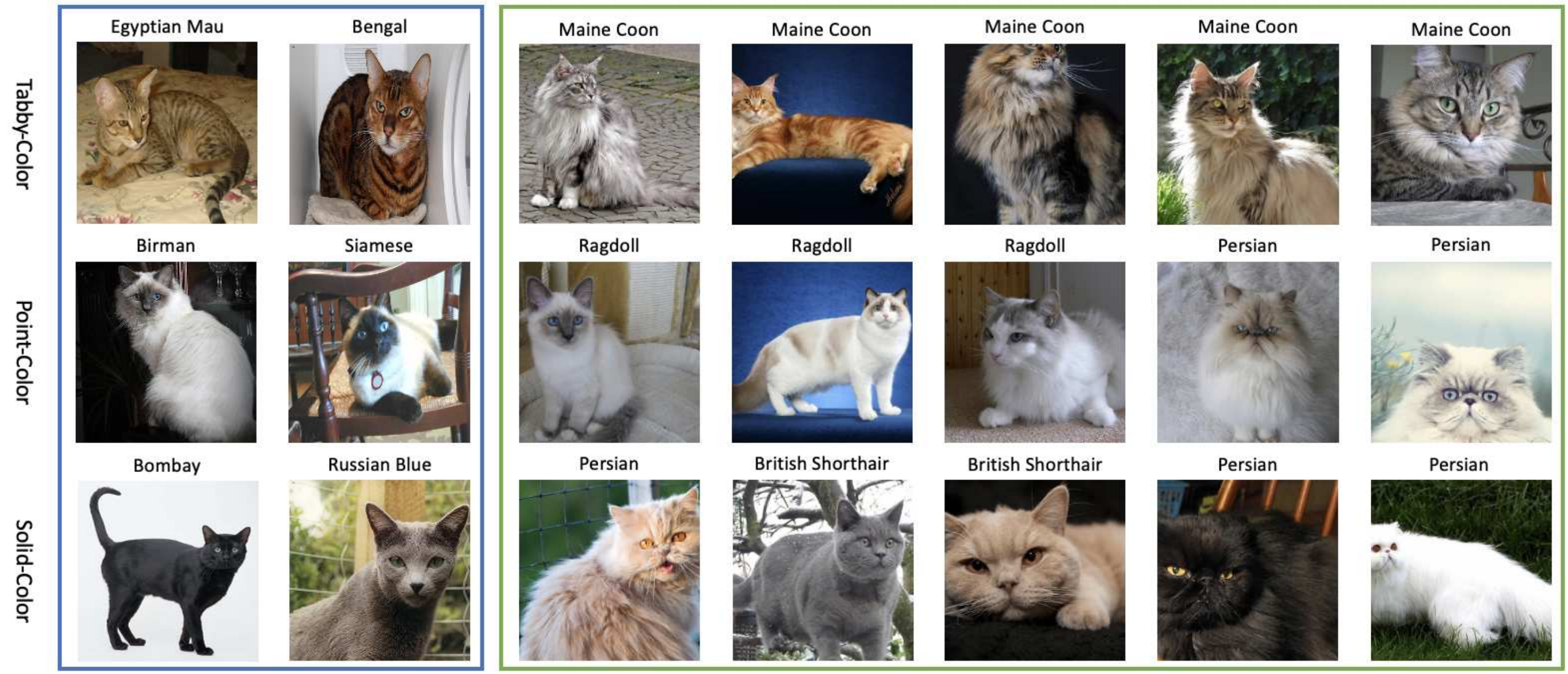}}
\caption{Interpretability results. The testing images at each row triggered the model inference process to traverse through the augmented node listed at far left (respectively for each row). The blue and green rectangles indicate deterministic and nondeterministic path prediction respectively.}
\label{fig:property}
\end{center}
\vskip -0.1in
\end{figure*}

As we mentioned in the Introduction section, an appealing by-product we characterize our framework with is much-enhanced interpretability than a black-box end-to-end system.
It can be perceived that the augmented label graph offers a ``decision process'' of our model, when performing an inference pass.
In Figure~\ref{fig:property} we plot three augmented nodes ($\langle$\emph{Tabby-Color}$\rangle$, $\langle$\emph{Point-Color}$\rangle$ and $\langle$\emph{Solid-Color}$\rangle$) with some of the corresponding images. 
More specifically, when conducting an inference pass on displayed images on each row, the decoded path will traverse through the listed augmented (feature) node on the far left.

Furthermore, the color of the blocks indicates the property of the predicted path obtained during inference on the test set.
The blue rectangle indicates the prediction is accomplished on deterministic paths while the green ones are nondeterministic paths.
This result indicates the validity of our training scheme involving policy gradient and normal sequence training techniques.

\subsection{Path correctness evaluation}
To further scrutinize the trained model, we manually check the predicted paths' correctness.
In particular we look into the predictions of nondeterministic paths, because for these paths, there aren't groundtruth associated with the ambiguous nodes, and sampling and policy gradient introduce a certain amount of uncertainty.

We center our inspection on the node $\langle$\emph{Tabby-Color}$\rangle$. This node appears on the deterministic paths leading to label nodes including $\langle$\emph{Abyssinian}$\rangle$, $\langle$\emph{Egyptian Mau}$\rangle$ and $\langle$\emph{Bengal}$\rangle$, and it turns out to be nondeterministic for $\langle$\emph{Maine Coon}$\rangle$ and $\langle$\emph{British-Shorthair}$\rangle$.
Among the predicted nondeterministic paths traversing through $\langle$\emph{Tabby-Color}$\rangle$ node, more than \textbf{90\%} of these samples' color pattern does conform to be a tabby color.

To us, this is a very exciting result because it shows that our model is indeed capable of performing some reasoning on the label graph and resolving the ambiguity existing on the nondeterministic paths.
We hope to leave the exploration of this line to the future work.

% \begin{figure}[t]
% \vskip 0.1in
% \begin{center}
% \centerline{\includegraphics[width=\columnwidth]{PropertyPath.eps}}
% \caption{Blue nodes and dash lines are part of cat-dog label graph. And the orange and yellow line represent the paths to corresponding samples predicted by model.(For samples of Maine Coon, the optional intermediate nodes set contains \emph{CSolidColor} and \emph{CTabby})}
% \label{fig:property2}
% \end{center}
% \vskip -0.1in
% \end{figure}

\subsection{Ablation study}
\label{subsec:ablation}
In this section, we attempt to identify the key factors influencing the model performance. 
We conduct a series of ablation study on the following factors: 
\begin{itemize}
    \item \textbf{the size of the label graph}. We compare our best result with the same architectural setup built on trimmed label graphs. Specifically, we obtain a \emph{medium}-size label graph by trimming down the augmented nodes by 36\%, and likewise, a \emph{small}-size label graph by trimming down the augmented nodes by 63\%.
    \item \textbf{minibatch construction}. In our training paradigm, each sample may correspond to several possible groundtruth paths. In practice, we could choose to sample (at most) $N_p$ target paths, and implement the gradient with a \emph{mean} or \emph{sum} pooling operator towards each input sample. We could also just sample a \emph{random} target path to do the training. Experiments show that the \emph{mean} operation performs the best (in Table~\ref{tab:baseline}).
\end{itemize}
The results are obtained from the Pet dataset group, displayed in Table~\ref{tab:ablation}.

\begin{table}[t]
\caption{Results of the ablation study. See text in section ~\ref{subsec:ablation} for details. The table shows the accuracy difference (\%) from the best results in Table~\ref{tab:baseline}.(\emph{OXF-PET} is short for \emph{Oxford-IIIT Pet}.)}
\label{tab:ablation} 
\vskip 0.15in
\setlength{\tabcolsep}{3pt}
\begin{center}
\begin{small}
\begin{sc}
\begin{tabular}{llrr}
\toprule
 & & Oxf-Pet & PetFusion \\
Graph size & & & \\
\midrule
& Medium graph & -0.14 & - \\
& Small graph & -0.19 & -0.1 \\
\midrule
Minibatch & & & \\
construction & & & \\
\midrule
& Sum & -0.71 & $-$ \\
& Random & -0.36 & $-$ \\
% \multirow{3}{*}{Graph Size} & \textbf{Label Graph} &  & \textit{\textbf{Oxf-Pet}} & \textit{\textbf{PetFusion}}\\ \cmidrule{2-5}
% & \textit{smaller Graph}  & & -0.19   & -0.1    \\  
% %& \textit{Extended Graph}  & & 94.66   & 94.95    \\ 
% \midrule
% \multirow{2}{*}{Minibatch} &\textbf{Dataset}  & \textit{\textbf{Mean}} & \textit{\textbf{Sum}} & \textit{\textbf{Random}} \\ \cmidrule{2-5}
% & \textit{Oxf-Pet}  & 94.66   & 93.95   & 94.30  \\ 
% \midrule
% \multirow{4}{*}{Train Strategy} & \textbf{Strategy}  &  & \textit{\textbf{Indep}} & \textit{\textbf{Toget}}\\ \cmidrule{2-5}
% & \textit{BatchB}  &  & 90.45 & 93.10     \\  
% & \textit{EpochB}  &  & 89.80 & 93.30     \\ 
% & \textit{NoB}  &   & 91.26 & 93.95       \\ 
\bottomrule
\end{tabular}
\end{sc}
\end{small}
\end{center}
\vskip -0.1in
\end{table}

\section{Outlooks}
\label{sec:conclusion}

%We explore the problem of text classification, which has a wide range of applications in our daily life. 
% We propose a framework BEIG that adopts the state-of-the-art language pre-training model BERT to perform text classification, and improves the quality of text classification by using the feature attribution method IG. Unlike the traditional end-to-end deep learning systems, BEIG does an iterative loop. It i) first classifies the data using BERT, ii) interprets the class predictions using IG to calculate the attribution of each token in an input sentence, iii) generates the perturbed input for each original input by randomly dropping out the tokens having negative contributions to predictions, and finally (iv) designs a mixed objective function to regularize the obtained model by taking both the original inputs and the perturbed inputs. Extensive experiments with a variety of public datasets evaluate the overall performance and different parts (i.e., the mixed objective function and the drop-out strategy) of BEIG, offering evidence that BEIG achieves the state-of-the-art performance without any additional knowledge. In the future, it is of interest to boost the efficiency of our framework by avoiding an iterative loop. 

Why train your neural network using just one dataset?

In this article, we study the problem of dataset joining, more specifically in label set joining when there is a labeling system discrepancy.
We propose a novel framework tackling this problem involving label space augmentation, recurrent neural network, sequence training and policy gradient. 
The trained model exhibits promising results both in performance and interpretability.
Furthermore, we also tend to position our work as a preliminary attempt to incorporate the rich domain knowledge (formatted as knowledge graphs of the labels) to boost the connectivism (such as a neural network classifier).
At last, we hope to use this work to motivate research for the multi-dataset joining setup for different tasks, and knowledge-driven label graphs with higher efficiency to be incorporated into deep learning.

\section*{Acknowledgement}
Jake Zhao, Wu Sai and Chen Gang are supported by the Key R\&D Program of Zhejiang Province (Grant No. 2020C01024).
JZ also thank Yiming Zhang for completing the preliminary baseline coding work for the ArXiv experiment.

\bibliography{paper}
\bibliographystyle{icml2021}

\clearpage
\end{document}